\documentclass[a4paper, times, 10pt,twocolumn]{article}
\usepackage[top=4.9cm,bottom=3.7cm,left=1.5cm,right=1.5cm]{geometry}
\usepackage{ICMLC}
\usepackage{times}
\usepackage{graphicx}
\usepackage{indentfirst}
\usepackage{latexsym}
\usepackage{booktabs}
\usepackage{cite}
\usepackage{xcolor}

\usepackage{amsthm, amsfonts, amsmath}

\newtheorem{definition}{Definition}
\newtheorem{example}{Example}

\newtheorem{proposition}{Proposition}

\newtheorem{corollary}{Corollary}
\newtheorem{remark}{Remark}

\providecommand{\Id}{\mathrm{Id}}
\providecommand{\dd}{\mathrm{d}}
\providecommand{\exp}{\mathrm{exp}}
\providecommand{\lie}{\mathfrak}
\providecommand{\R}{\mathbb{R}}

\providecommand{\GL}{\mathrm{GL}}
\providecommand{\gl}{\mathfrak{gl}}

\providecommand{\On}{\mathrm{O}}

\newcommand{\argmax}{\mathop{\mathrm{argmax}}}

\usepackage[margin=8pt,font=footnotesize,labelfont=bf,labelsep=period
]{caption}

\pdfpagewidth=\paperwidth
\pdfpageheight=\paperheight
\begin{document}

\title{ALGEBRAIC ADVERSARIAL ATTACKS ON INTEGRATED GRADIENTS}  

\author{\bf{\normalsize{LACHLAN SIMPSON${^1}$, FEDERICO COSTANZA${^2}$, KYLE MILLAR${^3}$, ADRIEL CHENG$^{1,3}$,}}\\ \bf{\normalsize{CHENG-CHEW LIM$^{1}$, AND HONG GUNN CHEW$^{1}$}}\\ 
\\
\normalsize{$^1$School of Electrical and Mechanical Engineering, The University of Adelaide, Australia}\\
\normalsize{$^2$School of Computer and Mathematical Sciences, The University of Adelaide, Australia} \\
\normalsize{$^3$Information Sciences Division, Defence Science \& Technology Group, Australia}
\\
\normalsize{E-MAIL: \{lachlan.simpson, estaban.costanza, adriel.cheng, cheng.lim, honggunn.chew\}@adelaide.edu.au}$^{1,2}$\\
\normalsize{\{kyle.millar1, adriel.cheng\}@defence.gov.au$^{3}$}
}

\maketitle \thispagestyle{empty}

\begin{abstract}{
   Adversarial attacks on explainability models have drastic consequences when explanations are used to understand the reasoning of neural networks in safety critical systems. Path methods are one such class of attribution methods susceptible to adversarial attacks.
   Adversarial learning is typically phrased as a constrained optimisation problem. In this work, we propose algebraic adversarial examples and study the conditions under which one can generate adversarial examples for integrated gradients. Algebraic adversarial examples provide a mathematically tractable approach to adversarial examples.
   }
   
\end{abstract}

\section{Introduction}
Deep learning provides state-of-the-art solutions to a wide array of classification tasks \cite{yolo}. A fundamental problem of deep learning is the black-box nature of the models \cite{Sejnowski_2020}. Gradient base-line attribution models address the black-box problem by providing an attribution of the input features to the prediction of neural network under analysis \cite{simpson2024probabilistic}. Several gradient explainability methods exist with the underlying assumption that analysis of the model's gradient highlights features with greatest impact on a prediction \cite{smilkov2017smoothgrad,pmlr-v70-sundararajan17a}.

Explainability models are prevalent in critical systems such as: cancer screening and cyber security \cite{ieeeXAIsurvey, Zhang_2022}. Reliable explanations are therefore as fundamental as a correct classification. Adversarial learning on neural networks has highlighted the drastic consequences a compromised neural network can have \cite{cnn_attack}. Likewise, adversarial learning on explainability models provides orthogonal reasoning used by the neural network to make a classification \cite{xai_attack}. Adversarial explainability is typically phrased as a constrained optimisation problem \cite{{Ghorbani_Abid_Zou_2019,NEURIPS2019_bb836c01,adv_perturb,BANIECKI2024102303}}. In this theoretical work, we take an algebraic approach to adversarial explainability. Lie groups and algebras are fundamental to the study of symmetry and reveals fundamental properties of functions. In this work, Lie theory is applied to study the conditions under which one can exploit algebraic properties of a neural network to produce adversarial explanations.

The contributions of this work is twofold:
\begin{enumerate}
    \item We provide a group theoretic description for adversarial explainability. Our approach provides a mathematically tractable approach of how adversarial examples are achieved.
    \item We demonstrate how to use Lie theory to compute algebraic adversarial examples for integrated gradients.
\end{enumerate}

The remainder of this work is structured as follows: Section 2 introduces related work on adversarial explainability, Section 3 provides the background on path attribution models. In Section 4 we provide an algebraic description of the symmetries of neural networks. In Section 5 we discuss how to compute the group of symmetries via the exponential map of the Lie algebra. In Section 6 we introduce the notion of algebraic adversarial examples. Furthermore, we theoretically demonstrate that four common base-point choices are susceptible to algebraic adversarial examples. We conclude in Section 7 with a discussion for future work. We assume familiarity with basic group and Lie theory. We refer the reader to \cite{ab_algebra_gallian} and \cite{Hall} respectively, for introductions. 
\section{Adversarial Attacks on Explainability Models}
Adversarial attacks cover a broad range of techniques. For our purposes, we consider the adversarial example. 
\begin{definition}
    Given an input $x$, a neural network $F$, an error tolerance $\varepsilon > 0$ and an explainability model $\phi$. An adversarial example $\tilde{x}$ satisfies the following conditions:
    \begin{enumerate}
        \item $\|\tilde{x} - x\|_{2} \leq \varepsilon$
        \item $F(\tilde{x}) = F(x)$
        \item $\phi(F,\tilde{x}) \neq \phi(F,x)$
    \end{enumerate}
\end{definition}

Ghorbani et al. \cite{Ghorbani_Abid_Zou_2019} generate adversarial explainability examples by perturbing an input in the gradient direction of the explanation. Generating adversarial examples is phrased as an optimisation problem subject to constraints 1) and 2) of Definition 1.  

Dombrowski et al. \cite{NEURIPS2019_bb836c01}  take a different approach where an adversarial example is optimised to be a specific explanation. Dombrowski et al. \cite{NEURIPS2019_bb836c01}  further demonstrate that the vulnerability of an explainability model is linked to the large curvature of the output manifold of the neural network. 

The central theme of related work is to minimise a loss function to produce close by points with the same classification, but drastically altered explanations \cite{adv_perturb,BANIECKI2024102303}. Our work is the first to take an algebraic approach to the problem of adversarial explanations. We assume, as with other work, that one has access to the neural network under attack.

\section{Path Methods}
Path methods are a specific class of post hoc explainability methods. Given a closed interval $I := [a, b] \subset \R$, a path $\gamma \colon I \to \R^{n}$ and a unit length vector $v \in \R^{n}$, the component of a path method $A^{\gamma} : \R^{n} \times \R^{n} \times F(\R^{n},\R^{m}) \to \R^{n}$ in the direction of $v$ is defined as
\begin{equation}\label{def:BAM}
    A^{\gamma}_{v}(x,x',F) = \int_{a}^{b} \langle \nabla F(\gamma(t)), v \rangle \langle \gamma'(t), v \rangle \dd t.
\end{equation}
In this way, for a given orthonormal basis $\lbrace v_{1}, \dots, v_{n} \rbrace$ of $\R^{n}$, $A^{\gamma}$ can be expressed as
\begin{equation}\label{eqn:attribution_basis}
    A^{\gamma}(x,x',F) = \sum\limits_{i = 1}^{n} A^{\gamma}_{v_{i}}(x,x',F) v_{i}.
\end{equation}
Particularly, for the standard orthonormal basis $\lbrace e_{1}, \dots, e_{n} \rbrace$ of $\R^{n}$, we obtain the usual definition
\begin{equation}
    A^{\gamma}_{e_{i}}(x,x',F) = \int_{a}^{b} \frac{\partial F}{\partial x_{i}}(\gamma(t)) \frac{\partial \gamma_{i}}{\partial t}(t) \dd t.
\end{equation}
Integrated gradients \cite{pmlr-v70-sundararajan17a} is a prominent path method where $\gamma$ is taken to be the straight line between points $x,x' \in \R^{n}$. For any pair of points $x, x' \in \R^{n}$, a neural network $F \in F(\R^{n},\R^{m})$, and the standard orthonormal basis of $\R^{n}$ integrated gradients is defined as
\[
\mathrm{IG}(x, x', F) := (x - x') \odot \int_{0}^{1} \nabla F(x' + t(x - x')) \dd t,
\]
where $\odot$ denotes the Hadamard product.

\section{The Symmetries of Neural Networks}
In this section we provide a group theoretic description of the symmetries of feed-forward neural networks. Fixing notation, let $\mathrm{M}_{n \times m}(\R)$ denote the real vector space of $n \times m$ matrices with coefficients in $\R$. The group of invertible linear transformations of $\R^{n}$, is the well known \textit{real general linear group}:
		\[
		\GL_{n}(\R) : = \lbrace g \in \gl_{n}(\R) \; : \; \det(g) \neq 0 \rbrace,
		\]
		where $\gl_{n}(\R)$ denotes the vector space of linear endomorphisms of $\R^{n}$, which shall be identified with the vector space of $n \times n$ matrices over the real numbers. We remark that $\GL_{n}(\R)$ is closed under transposition. Indeed, if $g^{t}$ denotes the transpose of $g \in \GL_{n}(\R)$, $\det(g^{t}) = \det(g) \neq 0$ and thus $g^{t} \in \GL_{n}(\R)$.
		
We consider the group action of $\GL_{n}(\R)$ on $F(\R^{n}, \R^{m})$, that is given by
		\[
		(g \cdot F)(x) = F(g^{t}x), \quad g \in \GL_{n}(\R), \quad F \in F(\R^{n}, \R^{m}).
		\]
        Even though we focus on the action defined above, for future reference, we consider $\R^{n}$ as a group with its usual addition as a group operation, we define the action of $\R^{n}$ on $F(\R^{n}, \R^{m})$ given by
        \[
        (u \cdot F)(x) = F(x - u), \quad u \in \R^{n}, \quad F \in F(\R^{n}, \R^{m}).
        \]

		\begin{definition}
			Let $F \in F(\R^{n}, \R^{m})$. A group $G$ will be called a group of symmetries of $F$, if $g \cdot F = F$ for all $g \in G$.
		\end{definition}
In what follows, we describe groups of symmetries contained in $\GL_{n}(\R)$, for a certain class of neural networks. Fixing notation, for a given activation function $f : \R^{d} \to \R^{m}$, a matrix $W \in \mathrm{M}_{d \times n}(\R)$, and a  bias $b \in \R^{d}$, the map 
		\begin{equation}\label{eqn:neural_network}
		F_{f, W, b} : \R^{n} \to \R^{m}, \quad x \mapsto f(Wx + b),
		\end{equation}
		will be a neural network, where $Wx$ is the composition of $k-1$ layers of a $k$-th layer neural network.
        \begin{proposition}
            Let $F_{f, W, b}$ be a neural network defined by equation (\ref{eqn:neural_network}). Then $\ker (W)$ is a group of symmetries of $F_{f, W, b}$.
        \end{proposition}    
        \begin{proof}
            For any $u \in \ker(W)$ we have
            \[
            (u \cdot F_{f, W, b})(x) = f(W(x-u) + b) = f(Wx + b) = F_{f, W, b}(x),
            \]
        as claimed.
        \end{proof}
        In the remainder of this work, $W$ will be considered as both a $d \times n$ matrix or the $d$-tuple of vectors $(w_{1}, \dots, w_{d})$ in $\R^{n}$, by identifying the $i$-th row of $W$ with $w_{i}$. Under this identification, we define the action of $\GL_{n}(\R)$ on $\mathrm{M}_{d \times n} (\R)$ given by
        \begin{equation}\label{def:action_W}
        g \cdot W = Wg^{t} = (gw_{1}, \dots, gw_{d}), \quad g \in \GL_{n}(\R).
        \end{equation}
        To describe the action of $\GL_{n}(\R)$ on the class of neural networks defined by equation (\ref{eqn:neural_network}), let us choose $g \in \GL_{n}(\R)$, and $F_{f, W, b}$ as above. It follows directly from the definitions of the action of $\GL_{n}(\R)$ on $F(\R^{n}, \R^{m})$, that
		  \begin{equation*}
		      (g \cdot F_{f, W, b})(x) = F_{f, W, b}(g^{t}x) = f(Wg^{t}x + b). 
		  \end{equation*}
            By equation (\ref{def:action_W}), the right-hand side of the above equation is nothing but $f((g\cdot W) x + b)$ and, therefore, we obtain
            \begin{equation}\label{eqn:action_neural_networks}
            g \cdot F_{f, W, b} = F_{f, g \cdot W, b}.
            \end{equation}
			\begin{proposition}
			Let $F_{f, W, b} : \R^{n} \to \R^{m}$ be a neural network defined by equation (\ref{eqn:neural_network}). Then 
            \begin{equation*}
             P_{W} : = \lbrace g \in \GL_{n}(\R) \; : \; g \cdot W= W \rbrace   
            \end{equation*}
            is a group of symmetries of $F_{f, W, b}$.
			\end{proposition}
            \begin{proof}
            From equation (\ref{eqn:action_neural_networks}), we observe that
            \[
            g \cdot F_{f, W, b} = F_{f, g \cdot W, b} = F_{f, W, b}.
            \]
            for all $g \in P_{W}$. Consequently, $P_{W}$ is a symmetry group of $F_{f, W, b}$.
            \end{proof}
            For convenience, $\mathcal{W}$ will denote the vector subspace of $\R^{n}$ spanned by $\lbrace w_{1}, \dots, w_{d} \rbrace$, i.e. the span of the rows of $W$.
            
            \begin{proposition}\label{prop:PW}
                Let $F_{f, W, b} : \R^{n} \to \R^{m}$ be a neural network defined by equation (\ref{eqn:neural_network}). Then 
            \begin{equation}\label{eqn:corollary_PW}
             P_{W} = \lbrace g \in \GL_{n}(\R) \; : \; gw= w, \; \forall w \in \mathcal{W} \rbrace.   
            \end{equation}
            \end{proposition}
			\begin{proof}
            It follows from equation (\ref{def:action_W}) and the fact that $\lbrace w_{1}, \dots, w_{d} \rbrace$ is a basis of $\mathcal{W}$.
			\end{proof}
			\begin{corollary}
                Let $F_{f, W, b} : \R^{n} \to \R^{m}$ be a neural network defined by equation (\ref{eqn:neural_network}). If $\dim \mathcal{W} = n$, $P_{W}$ is the trivial group, i.e. $P_{W} = \lbrace \Id \rbrace$.
			\end{corollary}
            \begin{proof}
            By Proposition \ref{prop:PW}, $P_{W}$ is comprised of the invertible $n \times n$ matrices that fix all vectors in $\mathcal{W}$. When $\dim \mathcal{W} = n$, $\mathcal{W}$ is equal to $\R^{n}$, and the only linear transformation fixing all vectors in $\R^{n}$ is identity matrix.
            \end{proof}
            The group of symmetries $P_{W}$, of a neural network such that $\dim \mathcal{W} = r < n$, can be easily computed in an appropriate basis of $\R^{n}$. Writing $\R^{n} = \mathcal{W} \oplus \mathcal{W}^{\perp}$, where $\mathcal{W}^{\perp}$ denotes the orthogonal complement of $\mathcal{W}$ in $\R^{n}$, and choosing an orthonormal basis $\lbrace e_{1}, \dots, e_{n} \rbrace$ of $\R^{n}$ such that $\lbrace e_{1}, \dots, e_{r} \rbrace$ is a basis of $\mathcal{W}$, any $g \in \gl_{n}(\R)$ will be represented in this basis by a matrix in block form
			\begin{equation}\label{eqn:block_matrix}
            g = 
        \begin{pmatrix}
		  A & B \\
		  C^{t} & D 
        \end{pmatrix},
			\end{equation}
        where $ A \in \gl_{r}(\R)$, $B, C \in \mathrm{M}_{r \times (n-r)}(\R)$ and $D \in \gl_{n-r}(\R)$.
        
        Particularly, when $g \in P_{W}$, we note from equation (\ref{eqn:block_matrix}) that $gw = w$ for all $w \in \mathcal{W}$ implies that $A = \Id$ and $C = 0$. Since $\det(g) \neq 0$ and $g$ is block upper-triangular, 
        \[
        \det(g) = \det(\Id)\det(D) = \det(D) \neq 0,
        \]
        hence $D \in \GL_{n-r}(\R)$. We have proved the following proposition.
        
        \begin{proposition}
            Let $F_{f, W, b} : \R^{n} \to \R^{m}$ be a neural network defined by equation (\ref{eqn:neural_network}), such that $\dim \mathcal{W} = r < n$. Then
			\begin{equation*}
			P_{W} \simeq
			\left\lbrace
			\begin{pmatrix}
			\Id & B \\
			0 & A
			\end{pmatrix}
			 : A \in \GL_{n-r}(\R), \; B \in \mathrm{M}_{r \times (n- r)}(\R) 
			\right\rbrace.
			\end{equation*}
        \end{proposition}

\section{The Lie Algebra of Neural Network Symmetries}
In this section, we discuss how to generate elements of the group of symmetries defined in Section 4 via the exponential of matrices. Recall that the exponential of matrices is the map $\exp : \gl_{n}(\R) \to \GL_{n}(\R)$, defined by
            \begin{equation}\label{def:exp}
 		\exp(A) := \sum\limits_{k = 0}^{\infty} \frac{A^{k}}{k!} = \frac{1}{0!}\Id + \frac{1}{1!}A + \frac{1}{2!}A^{2} + \dots
 		\end{equation}

            It is well-known that the Lie algebra $\lie{g}$, of a matrix Lie group $G \subseteq \GL_{n}(\R)$, consists of all $A \in \gl_{n}(\R)$ such that $\exp(tA) \in G$ for all $t \in \R$, and we remark that image of $\exp$ on $\gl_{n}(\R)$ is $\GL^{+}_{n}(\R) : = \lbrace g \in \GL_{n}(\R) \; : \; \det(g) > 0 \rbrace.$
            For more details we refer to \cite[Section 3.3]{Hall}.
            

            
            We denote the Lie algebra of $P_{W}$ by $\lie{p}_{W}$, and define the set $P_{W}^{+} : = P_{W} \cap \GL_{n}^{+}(\R) = \exp(\lie{p}_{W})$.
            
            \begin{remark}
                $\lie{p}_{W}$ is the Lie algebra of both, $P_{W}$ and $P_{W}^{+}$.
            \end{remark}      
			We will proceed to describe an algorithm to compute elements of $P_{W}^{+}$, explicitly, without making use of any specific basis of $\R^{n}$.
            \begin{proposition}\label{prop:lie_algebra}
                The Lie algebra of $P_{W}^{+}$ is 
                \[
                \lie{p}_{W} = \lbrace A \in \gl_{n}(\R) \; : \; Aw = 0, \; \forall w \in \mathcal{W} \rbrace.
                \]
            \end{proposition}
            \begin{proof}
            For any $A \in \lie{p}_{W}$, $\exp(tA) \in P_{W}$ for all $t \in \R$. Then, $\exp(tA) w = w$ for all $w \in \mathcal{W}$ and
            \[
            0 = \frac{d}{dt} \; \exp(tA)w = \exp(tA)Aw, \quad \forall t\in \R.
            \]
            Particularly, when $t = 0$, the above equation becomes $Aw = 0$, since $\exp(0) = \Id$.
            \end{proof}
            
            To compute $\lie{p}_{W}$ explicitly, we define $A_{xy} \in \gl_{n}(\R)$, given by 
            \[
            A_{xy}z = xy^{t}z = \langle y, z \rangle x, \quad x, y, z \in \R^{n}.
            \]
            The kernel of $A_{xy}$ is precisely the vector space orthogonal to $\R y := \lbrace  k y \colon k \in \R \rbrace$. From this construction it is not difficult to see that $\mathfrak{p}_{W} = \mathrm{span} \lbrace A_{xy} \; : \; y \in \mathcal{W}^{\perp} \rbrace$. The following proposition is a consequence of this construction and Proposition \ref{prop:lie_algebra}.
            \begin{proposition}\label{prop: PW_generated}
                $P^{+}_{W}$ is generated by 
                \[
                \left\lbrace \exp\left( \sum_{i = 1}^{k} A_{x_{i}y_{i}} \right) \; : \; y_{i} \in \mathcal{W}^{\perp}, \; k \geq 1 \right\rbrace .
                \]
            \end{proposition}
            Proposition 6 thereby provides an explicit way to compute elements of the group of symmetries of the neural network.

\section{Algebraic Attacks on Integrated Gradients}
In this section, we define the notion of algebraic adversarial examples. We demonstrate that the symmetry groups defined in Section 4 generate adversarial examples for integrated gradients. Further, we demonstrate how to compute adversarial examples for several common base-line choices. The cases $G \subseteq \GL_{n}(\R)$ and $G \subseteq \R^{n}$ will be considered separately.

\begin{definition}
    Let $F \in F(\R^{n}, \R^{m})$ be a neural network, $G$ a group of symmetries of $F$ contained in $\GL_{n}(\R)$ or $\R^{n}$, $\phi$ an explainablity model and $x$ an input. We say that $\tilde{x}$ is an algebraic adversarial example if it is an adversarial example satisfying Definition 1 and
    \begin{enumerate}
    \item $\tilde{x} = g^{t} x$ for some $g \in G \subseteq \GL_{n}(\R)$, or
    \item $\tilde{x} = x - u$ for some $u \in G \subseteq \R^{n}$.
    \end{enumerate}
\end{definition}

To construct algebraic adversarial examples for integrated gradients, we address each condition listed in Definition~1. Condition 1 requires that the difference between the adversarial point $\tilde{x}$ and the ``clean" point $x$ satisfy the inequality 
\begin{equation}
        \| \tilde{x} - x \| \leq \varepsilon.
\end{equation}
for a given error threshold $\varepsilon \geq 0$. To address this problem for algebraic adversarial examples, we prove the proposition below and remark that the result is valid for any explainability model.


\begin{proposition}
        Let $F$ be a neural network defined by equation (\ref{eqn:neural_network}), $G$ a group of symmetries of $F$, $x \in \R^{n}$ and $\varepsilon > 0$. Then $\tilde{x}$ given by
        \begin{enumerate}
        \item $\tilde{x} = g^{t}x$, when $G \subseteq \GL_{n}(\R)$, or
        \item $\tilde{x} = x + u$, when $G \subseteq \R^{n}$
        \end{enumerate}
        satisfies equation (10).
    \end{proposition}

    \begin{proof}
         Suppose $G \subseteq \GL_{n}(\R)$ and let $\tilde{x} = \exp(kA)x$, with $A \in \lie{g}$ and $k \in \R$. It follows from the triangle inequality that choosing $k$ such that 
        \begin{equation*}
            |k| \leq \frac{1}{\|A \|}\log \left (\frac{\varepsilon}{\|x\|} + 1 \right),
        \end{equation*}
        $\| \tilde{x} - x \|$ will satisfy the error threshold. In the case when $G \subseteq \R^{n}$, it is enough to choose $u$ such that $\|u\| \leq \varepsilon$.
    \end{proof}

    Proposition 7 provides us with candidates for algebraic adversarial examples, which will satisfy condition 2 by construction, since $g \cdot F = F$ for all $g \in G$, by definition.

    It is only left to address condition 3. In this case, we restrict to integrated gradients. Before proceeding recall that the \textit{orthogonal group} is the subgroup of $\GL_{n}(\R)$ given by
    \[
    \mathrm{O}(n) = \lbrace g \in \GL_{n}(\R) \; : \; g^{t} = g^{-1} \rbrace,
    \]
    and we remark that since it is a group, for any $g \in \On(n)$, we have that $g^{t} = g^{-1} \in \On(n)$. 
    \begin{proposition}
        Let $g \in \On(n)$ and $u, v \in \R^{n}$, with $\|v \|_{2} = 1$. Then
        \begin{equation}\label{ig_on_sym1}
            \mathrm{IG}_{v}(gx, gx', g \cdot F) = \mathrm{IG}_{g^{t}v}(x, x', F),
        \end{equation}
        \begin{equation}\label{ig_t_sym1}
            \mathrm{IG}_{v}(x + u, x' + u, u \cdot F) = \mathrm{IG}_{v}(x, x', F).
        \end{equation}
    \end{proposition}
    \begin{proof}
        To begin, we prove that equation (\ref{ig_on_sym1}) holds. Let $\gamma_{x'}^{x}(t) = x' + t(x - x')$. It is straightforward to see that
        \begin{equation}\label{prop_eqn1}
        \gamma_{gx'}^{gx}(t) = g \gamma_{x'}^{x}(t), \quad (\gamma_{gx'}^{gx})'(t) = g (\gamma_{x'}^{x})'(t).
        \end{equation}
        Taking the inner product of the second of the above equations with $v$, we get
        \begin{equation}\label{prop_eqn2}
        \langle (\gamma_{gx'}^{gx})'(t), v \rangle = \langle g(\gamma_{x'}^{x})'(t), v \rangle = \langle (\gamma_{x'}^{x})'(t), g^{t}v \rangle.
        \end{equation}
        A simple application of the chain rule shows that
        \begin{equation}\label{prop_eqn3}
            \nabla(g \cdot F)(x) = g \; \nabla F (g^{t}x),    
        \end{equation}
        where $\nabla F (x)$ is considered as a column vector. From equations (\ref{prop_eqn1})  and (\ref{prop_eqn3}) it follows that
        \begin{equation}\label{prop_eqn4}
            \nabla(g \cdot F)( \gamma_{gx'}^{gx}(t)) = g \nabla F (g^{t} g \gamma_{x'}^{x}(t)).
        \end{equation}
        Since $g \in \On(n)$, we have that $g^{t}g = \Id$, and taking the inner product of equation (\ref{prop_eqn4}) with $v$ we obtain
        \begin{equation}\label{prop_eqn5}
            \langle \nabla(g \cdot F)( \gamma_{gx'}^{gx}(t)), v \rangle = \langle \nabla F (\gamma_{x'}^{x}(t)), g^{t}v \rangle.
        \end{equation}
        By replacing equations (\ref{prop_eqn2}) and (\ref{prop_eqn5}) in equation (\ref{def:BAM}), we have verified that equation (\ref{ig_on_sym1}) holds.

        The approach for the remaining case will be similar as the one above. Firstly, we note that
        \begin{equation}\label{prop_eqn6}
            \gamma_{x' + u}^{x + u}(t) = \gamma_{x'}^{x}(t) + u, \quad (\gamma_{x' + u}^{x + u})'(t) = (\gamma_{x'}^{x})'(t).
        \end{equation}
        Applying the chain rule, we get
        \begin{equation}\label{prop_eqn7}
            \nabla (u \cdot F)(x) = \nabla F (x - u).
        \end{equation}
        Then
        \begin{equation}\label{prop_eqn8}
            \nabla (u \cdot F)(\gamma_{x' + u}^{x + u}(t)) = \nabla F (\gamma_{x'}^{x}(t))
        \end{equation}
        follows from equations (\ref{prop_eqn6}) and (\ref{prop_eqn7}). Lastly, replacing equations (\ref{prop_eqn6}) and (\ref{prop_eqn8}), we show that equation (\ref{ig_t_sym1}) holds.
    \end{proof}

    \begin{remark}
    Proposition 8 demonstrates that the group action is natural in the sense that integrated gradients is preserved under rotations and translations to data. We employ Proposition 8 in the following section to compute adversarial examples for integrated gradients.
    \end{remark}

    We conclude this section by constructing algebraic adversarial examples for integrated gradients, in terms of the symmetry groups computed in Proposition 1 and Proposition 2. Our adversarial examples are constructed via Proposition 7, hence conditions~1 and 2  of Definition 1 are automatically satisfied.
    \begin{example}
        Taking $G = P_{W} \cap \On(n)$, we choose $\tilde{x} = g^{t}x$, for some $g \in G$. To verify that $\tilde{x}$ satisfies condition~3, we note that
        \[
        \mathrm{IG}_{v}(g^{t}x, x', F) = \mathrm{IG}_v(g^{-1}x, g^{-1}gx', g^{-1}g \cdot F),
        \]
        since $g^{t} = g^{-1}$. By Proposition 8, the right-hand side of the above equation is equal to $\mathrm{IG}_{g^{t}v}(x, gx', g \cdot F)$ and thus 
        \[
        \mathrm{IG}_{v}(g^{t}x, x', F) = \mathrm{IG}_{g^{t}v}(x, gx', g\cdot F) = \mathrm{IG}_{g^{t}v}(x, gx', F), 
        \]
        since $g \cdot F = F$ by assumption. Therefore
        \[
        \mathrm{IG}_{v}(g^{t}x, x', F) = \mathrm{IG}_{g^{t}v}(x, gx', F) \neq \mathrm{IG}_{v}(x, x', F).
        \]
    \end{example}
    The following example is built analogously.
    \begin{example}
        Taking $G = \ker(W)$, we choose $\tilde{x} = x - u$, with $u \in G$, given by Proposition 7. As before, we note that
        \[
        \mathrm{IG}_{v}(x - u, x', F) = \mathrm{IG}_{v}(x - u, (x' + u) - u, - u \cdot (u \cdot F)).
        \]
        By Proposition 8, the above equation becomes
        \[
        \mathrm{IG}_{v}(x - u, x', F) = \mathrm{IG}_{v}(x, x' + u, u \cdot F) = \mathrm{IG}_{v}(x, x' + u, F),
        \]
        where the last equality follows from $u \in G$. Therefore
        \[
        \mathrm{IG}_{v}(x - u, x', F) = \mathrm{IG}_{v}(x, x' + u, F) \neq \mathrm{IG}_{v}(x, x' , F).
        \]
    \end{example}
    Below are four common base-point choices \cite{sturmfels2020visualizing}.

\begin{enumerate}
    \item \textbf{Zero base-point}. Here the base-point for all points is a constant zero vector $ \alpha^{\text{zero}} = 0$.
    \item \textbf{Maximum Distance}. For a given input $x$, $\alpha$ is defined as the point of maximum distance from $x$.
    \begin{equation*}
        \alpha_{x}^{\max} = \argmax \|x-y\|_{p}.
    \end{equation*}
    Usually $p = 1$ or $2$.
    
    \item \textbf{Uniform}. We sample uniformly over a valid range of the dataset
    \begin{equation*}
    \alpha^{\text{uniform}}_{i} \sim U(\min_{i},\max_{i}).
    \end{equation*}

    \item \textbf{Gaussian}. A Gaussian filter is applied to the input x. 
    \begin{equation*}
        \alpha^{\text{Gaussian}} = \sigma \cdot v+x,
    \end{equation*}
    where $v_{i} \sim \mathcal{N}(0,1)$ and $\sigma \in \mathbb{R}$.
    We require the $\alpha^{\text{Gaussian}}$ is still within the data distribution so $\alpha^{\text{Gaussian}} \to \alpha^{\text{Uniform}}$ as $\sigma \to \infty$ \cite{sturmfels2020visualizing}. 
\end{enumerate}

Of this list of common base-point choices, we note that Example 2 provides algebraic adversarial examples for all of base-points in the aforementioned list of common choices. In fact, it provides adversarial examples for integrated gradients for any base-point choice $x' \in \R^{n}$. Example 1 fails to be an adversarial example when we choose the zero base-point $x' = 0$, however, it provides adversarial examples for any choice of base-point in $\R^{n}\backslash\{0\}$.

\section{Conclusions and Future Work}
In this work, we introduced the notion of algebraic adversarial examples for explainability models. We provided algebraic conditions under which one can generate adversarial examples for integrated gradients. Furthermore, we provided an algorithm to compute the generators of adversarial examples and demonstrated that these are adversarial examples for common base-point choices. In future work, we seek to experimentally validate our theoretical results and extend our work to other explainability models.

\section*{Acknowledgements}
The Commonwealth of Australia (represented by the Defence Science and Technology Group) supports this research through a Defence Science Partnerships agreement. Lachlan Simpson is supported by a scholarship from the University of Adelaide.

\bibliographystyle{IEEEtran}

\bibliography{main} 

\newpage

\end{document}